%% file: aaai2026.tex
\documentclass[letterpaper]{article} 
\usepackage{aaai2026}  
\usepackage{times}  
\usepackage{helvet}  
\usepackage{courier}  
\usepackage[hyphens]{url}  
\usepackage{graphicx} 
\usepackage{natbib}  
\usepackage{caption} 
\usepackage{algorithm}
\usepackage{algorithmic}
\usepackage{lipsum}
\usepackage{multirow} 
\usepackage{xcolor}
\usepackage{caption}
\usepackage{svg}
\usepackage{enumitem}
\usepackage{xspace}
\usepackage{tabularx}
\usepackage{makecell}

\usepackage{amsmath}
\usepackage{amssymb}

\usepackage{pifont}  

\usepackage{newfloat}
\usepackage{listings}
\DeclareCaptionStyle{ruled}{labelfont=normalfont,labelsep=colon,strut=off} 
\floatstyle{ruled}
\newfloat{listing}{tb}{lst}{}
\floatname{listing}{Listing}
\title{Unlocking the Potential of MLLMs in Referring Expression Segmentation \\ via a Light-weight Mask Decoder}
\author {
    Jingchao Wang\textsuperscript{\rm 1}\equalcontrib,
    Zhijian Wu\textsuperscript{\rm 3}\equalcontrib,
    Dingjiang Huang\textsuperscript{\rm 1}\thanks{Corresponding author},
    Yefeng Zheng\textsuperscript{\rm 3},
    Hong Wang\textsuperscript{\rm 2}$^{\dagger}$
}

\affiliations {
    \textsuperscript{\rm 1}School of Data Science and Engineering, East China Normal University\\
    \textsuperscript{\rm 2}School of Life Science and Technology, Xi'an Jiaotong University\\
    \textsuperscript{\rm 3}Medical Artificial Intelligence Laboratory, Westlake University\\
    jcwang@stu.ecnu.edu.cn, wuzhijian@westlake.edu.cn, \\ djhuang@dase.ecnu.edu.cn, zhengyefeng@westlake.edu.cn, hongwang01@xjtu.edu.cn
}

\usepackage{bibentry}

\begin{document}

\maketitle

\newcommand{\mymethod}{\textbf{MLLMSeg}}

\begin{abstract}
Reference Expression Segmentation (RES) aims to segment image regions specified by referring expressions and has become popular with the rise of multimodal large models (MLLMs). 
While MLLMs excel in semantic understanding, their token-generation paradigm struggles with pixel-level dense prediction. 
Existing RES methods either couple MLLMs with the parameter-heavy Segment Anything Model (SAM) with 632M network parameters or adopt SAM-free lightweight pipelines that sacrifice accuracy. 
To address the trade-off between performance and cost, we specifically propose {\mymethod}, a novel framework that fully exploits the inherent visual detail features encoded in the MLLM vision encoder without introducing an extra visual encoder. 
Besides, we propose a detail-enhanced and semantic-consistent feature fusion module (DSFF) that fully integrates the detail-related visual feature with the semantic-related feature output by the large language model (LLM) of MLLM. 
Finally, we establish a light-weight mask decoder with only 34M network parameters that optimally leverages detailed spatial features from the visual encoder and semantic features from the LLM to achieve precise mask prediction. 
Extensive experiments demonstrate that our method generally surpasses both SAM-based and SAM-free competitors, striking a better balance between performance and cost. Code is available at \url{https://github.com/jcwang0602/MLLMSeg}.
\end{abstract}


\section{Introduction}

\begin{figure}[!t] 
    \centering  
    \includegraphics[width=0.9\linewidth]{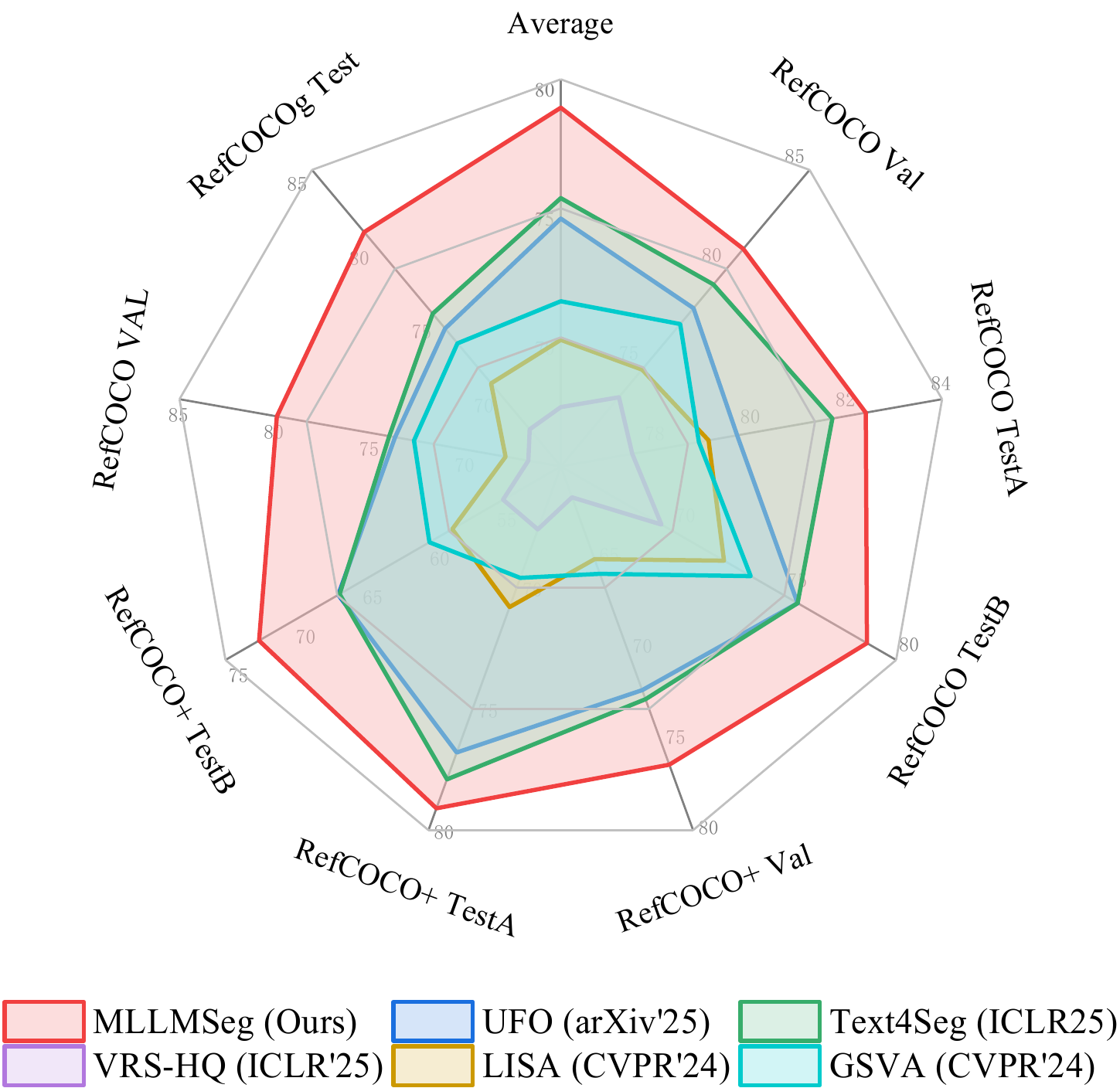} 
    \caption{Performance of {\mymethod} on Referring Expression Segmentation (RES), showing consistent improvement over state-of-the-art methods.}
    \label{fig:intro}  
\end{figure}


Referring Expression Segmentation (RES) lies at the intersection of computer vision and natural language processing~\cite{wang2025progressive}. Given an input image, the goal of the RES task is to segment an object or region designated by a natural language expression. 

Multimodal large language models (MLLMs) have recently demonstrated strong capabilities in understanding and reasoning across diverse modalities, particularly vision and language. Benefiting from large-scale pretraining and instruction tuning, MLLMs excel at various tasks, such as visual question answering, captioning, referring expression comprehension, and reasoning-based multimodal interaction. However, despite their strong capability in semantic understanding, MLLMs inherently operate in a token generation paradigm, which is well-suited for producing text sequences but not for generating dense and fine-grained pixel-level predictions~\cite{ren2024pixellm}.





Recently, in order to adapt MLLMs to the RES task for pixel-wise mask prediction, researchers have proposed two main paradigms: Segment Anything Model (SAM) \cite{kirillov2023segment}-based and SAM-free pipelines. 
Specifically, SAM-based methods generally utilize MLLMs to generate prompts, which are then fed to the SAM mask decoder for pixel-level prediction. For instance, LISA \cite{lai2024lisa} utilized MLLMs to generate prompt tokens for SAM mask generation, and GSVA \cite{xia2024gsva} introduced multiple $[SEG]$ tokens and a $[REJ]$ token to deal with the Generalized Referring Expression Segmentation (GRES) task for multi-object segmentation and empty target rejection. 
Although achieving promising performance, the introduction of SAM would inevitably bring many network parameters, causing a large storage cost (\emph{i.e.}, 632M). 
To alleviate this issue, the SAM-free pipeline has emerged. 
For example, in UFO \cite{tang2025ufo}, the authors reformulated the segmentation procedure as embedding retrieval by calculating the dot-product similarity between mask tokens and image features extracted by MLLMs. Text4Seg \cite{text4seg} introduced a new text-as-mask paradigm that casts image segmentation as a text generation problem, eliminating the need for additional decoders. Although these SAM-free approaches avoid introducing additional network parameters, the insufficient feature propagation generally causes limited performance improvement (refer to Fig.~\ref{fig:intro}).

To strike a better balance between performance and cost, in this paper, we specifically propose a novel approach, called {\mymethod}, in order to finely unleash the potential of MLLMs in addressing the RES task. Specifically, we carefully analyze that the visual feature extracted by the visual encoder of MLLM inherently contains rich detail information, which are useful for the fine-grained segmentation prediction. Inspired by this analysis, we propose to directly utilize the shallow detail-related visual feature without introducing an extra visual encoder, like SAM visual encoder in LISA and GSVA. Then we construct a detail enhancement and semantically consistent feature fusion module (DSFF)
as well as a flexible upsampling mechanism, which fully integrates the detail-related visual feature with the semantic-related feature output by the large language model (LLM) of MLLM. Finally, we establish a light-weight mask decoder with only 34M network parameters, which optimally leverages detailed spatial features from the visual encoder and semantic features from the LLM to achieve precise mask prediction. 
As shown in Figure~\ref{fig:intro}, our {\mymethod} achieves comprehensive performance superiority over contemporary methods under identical training conditions, achieving a better trade-off between performance and cost. Our main contributions are three-fold:
\begin{itemize}
    \item We specifically propose a novel framework, called {\mymethod}, for adapting MLLM to deal with the referring expression segmentation task. The main characteristic of our proposed method lies in that it does not require the introduction of a large-scale, pre-trained basic segmentation model like SAM, but only uses a light-weight mask decoder to achieve comparable performance.
    \item We propose a \textbf{D}etail-enhanced and \textbf{S}emantically consistent \textbf{F}eature \textbf{F}usion module (\textbf{DSFF}), that can efficiently inject the shallow detail-related visual feature extracted by the visual encoder of MLLM into the deep semantic-related feature extracted by the LLM of MLLM, and then boost the accurate mask prediction.
    \item Comprehensive experimental results demonstrate that our proposed {\mymethod} can always achieve the average SOTA performance across multiple tasks, including Referring Expression Segmentation (RES), Referring Expression Comprehension (REC), and Generalized Referring Expression Segmentation (GRES)

\end{itemize}

\section{Related Work}

\subsection{Referring Expression Segmentation}
Referring Expression Segmentation (RES) is a popular task at the intersection of computer vision and natural language processing, aiming at segmenting the corresponding target object or region in an image based on a given natural language description (Referring Expression). 
CEFNet\cite{cefnet} proposes an encoder fusion network that transforms the visual encoder into a multimodal feature learning network and utilizes language to progressively refine these multimodal features. 
LST\cite{lts} tackles referring image segmentation by decoupling it into referring object prior prediction and fine mask generation, achieving superior performance through explicit position prior modeling. 
In order to improve cross-modal alignment, LAVT\cite{yang2022lavt} introducing early fusion of linguistic and visual features within the intermediate layers of a vision Transformer encoder, enabling stronger multi-modal context modeling and more accurate segmentation without relying on heavy cross-modal decoders.


\subsection{Generalized Referring Expression Segmentation}
Referring Expression Segmentation (RES) has garnered significant attention in recent years. However, most traditional approaches impose strong pre-defined constraints on the task, often excluding expressions that refer to no target or multiple targets. To address these limitations, Chang Liu et al.\cite{liu2023gres} introduced a novel task, Generalized Referring Expression Segmentation (GRES), along with a large-scale benchmark dataset, gRefCOCO. They also proposed a baseline method, ReLA, which explicitly models the relationships between different image regions and linguistic components.
Leveraging the strong inference capabilities of MLLMs, GSVA\cite{xia2024gsva} introduces multiple $[SEG]$ tokens and a novel $[REJ]$ token, achieving effective multi-object segmentation as well as the rejection of empty targets.

\begin{figure*}[h] 
    \centering  
    \includegraphics[width=0.9\linewidth]{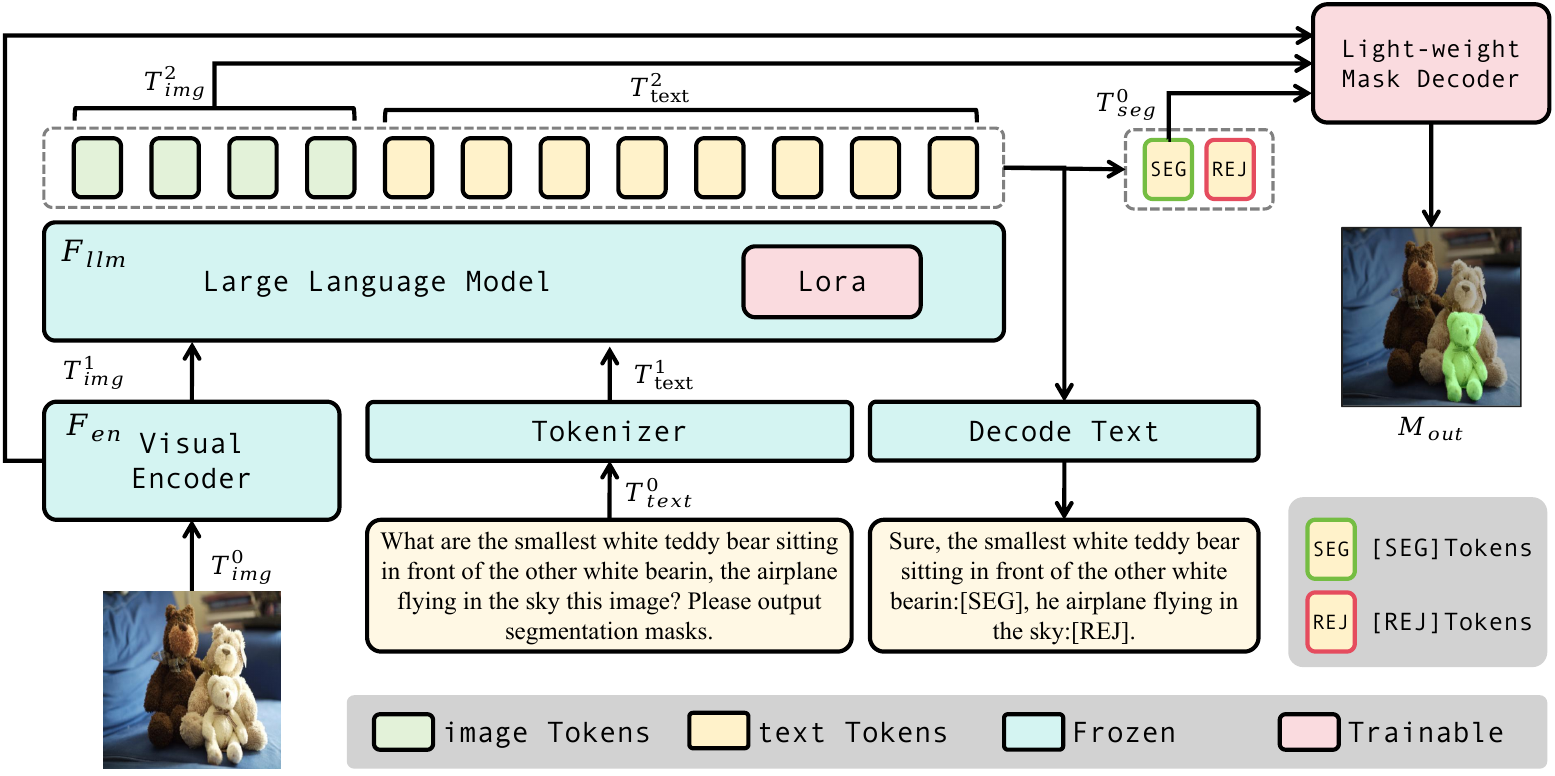} 
    \caption{Overview of the proposed {\mymethod}, where the detailed structure of light-weight mask decoder is shown in Figure~\ref{fig:dsff2}.}
    \label{fig:method}  
\end{figure*}

\subsection{Multi-modal Large Language Models}


MLLMs \cite{zhang2025pixel} have made significant progress by aligning powerful visual encoders with Large Language Models (LLMs). 
However, most existing MLLMs are limited to text-only outputs, which restricts their applicability in more diverse and complex multimodal scenarios.
LISA\cite{lai2024lisa} introduces SEG tokens to leverage the reasoning abilities of MLLMs and address the challenge of segmentation ambiguity. 
However, LISA struggles with the unique challenges posed by Generalized Referring Expression Segmentation (GRES), primarily due to the rigidity of its SEG token design. 
To overcome this limitation, GSVA\cite{xia2024gsva} proposes a more flexible approach by introducing multiple SEG tokens along with a novel REJ token, enabling effective multi-object segmentation and the rejection of expressions with no corresponding target.
While these methods achieve impressive results with powerful SAMs, they also incur significant parameters.


\section{Method}
In this section, we first present an overview of our proposed {\mymethod} and then describe the involved lightweight mask decoder for enabling MLLMs to accomplish the efficient segmentation. Finally, we introduce the training objective for optimizing the model in an end-to-end manner.

\subsection{Model Architecture}
Figure \ref{fig:method} illustrates the entire network architecture of our proposed {\mymethod}, which mainly consists of a multimodal large language model (MLLM) and a light-weight mask decoder. Here the MLLM is composed of a vision encoder and a large language model, which is utilized to extract visual features and generate proxy tokens as $[SEG]$ and $[REJ]$. The proposed light-weight mask decoder is used to generate segmentation masks based on visual features and proxy tokens. Please note that we follow GSVA~\cite{xia2024gsva} and adopt multiple $[SEG]$ tokens for multi-target segmentation and a $[REJ]$ token for empty target rejection. However, unlike GSVA which is built on the heavy-weight SAM, we carefully design a light-weight masked decoder which is jointly trained with multi-task language models (MLLMs) to achieve the accurate segmentation prediction.



\subsubsection{Multimodal Large Language Model.} 
Given an input image $T^{0}_{img} \in \mathbb{R}^{448 \times 448 \times 3}$ and the text description $T^{0}_{text}$ that specifies the target to be segmented, the MLLM first feeds them to a visual encoder $F_{en}$ and a tokenizer, respectively, expressed as:
\begin{equation}
\begin{split}
    &T^{1}_{img} = \text{Reshape}(F_{en}(T^{0}_{img})), \\
    &T^{1}_{text} = \text{Tokenizer}(T^{0}_{text}),
    \end{split}
\end{equation}
where $\text{Reshape}(\cdot)$ means converting the 2-dimensional spatial dimensions of the visual features into a one-dimensional sequence, which is consistent with the shape of the text sequence. $T^{1}_{img}$ and  $T^{1}_{text}$ are the detail-related visual tokens embedding and the text tokens, respectively.



Taking the concatenation of the visual token $T^{1}_{img}$, and the text token $T^{1}_{text}$ as the input sequence $T^{llm}_{input}$, and then feeding it into the large language model (LLM) $F_{llm}$, through the auto-regessive processing, we can get the final output token $T^{llm}_{output}$. The process can be formulated as:
\begin{align}
    & T^{llm}_{input} =  \text{Concat}(T^{1}_{img}, T^{1}_{text}), \\
    & T^{llm}_{output} = F_{llm}(T^{llm}_{input}), \\
    & [T^{2}_{img}, T^{2}_{text}]= \text{Split}(T^{llm}_{output}),
\end{align}
where $\text{Concat}(\cdot)$ denotes the concatenation procedure along the length dimension of the token sequence and $\text{Split}(\cdot)$ denotes dividing the output $T^{llm}_{output}$ of the MLLM into two parts for the subsequent processing, 
\emph{i.e.}, the semantic-related visual token $T^{2}_{img}$ and the semantic-related text token $T^{2}_{text}$. 
$T^{2}_{text}$ contains several $[SEG]$ Tokens and $[REJ]$ Tokens.
Following GSVA, several special tokens, $[SEG]$ and $[REJ]$, are appended to the vocabulary to provide MLLM segmentation and identify empty object capabilities. 
MLLM learns to predict the $[SEG]$ in the output sequence to represent the target to be segmented and $[REJ]$ to represent the target specified in the user instruction that is not in the image.

After the processing by MLLM, the image token $T^{1}_{img}$ extracted by the visual encoder $F_{en}$, the image token $T^{2}_{img}$ output by the LLM $F_{llm}$, and the special token $[SEG]$ $T_{seg}$ are fed into the subsequent light-weight mask decoder for segmentation prediction.

\begin{figure}[t] 
    \centering  
    \includegraphics[width=1.0\linewidth]{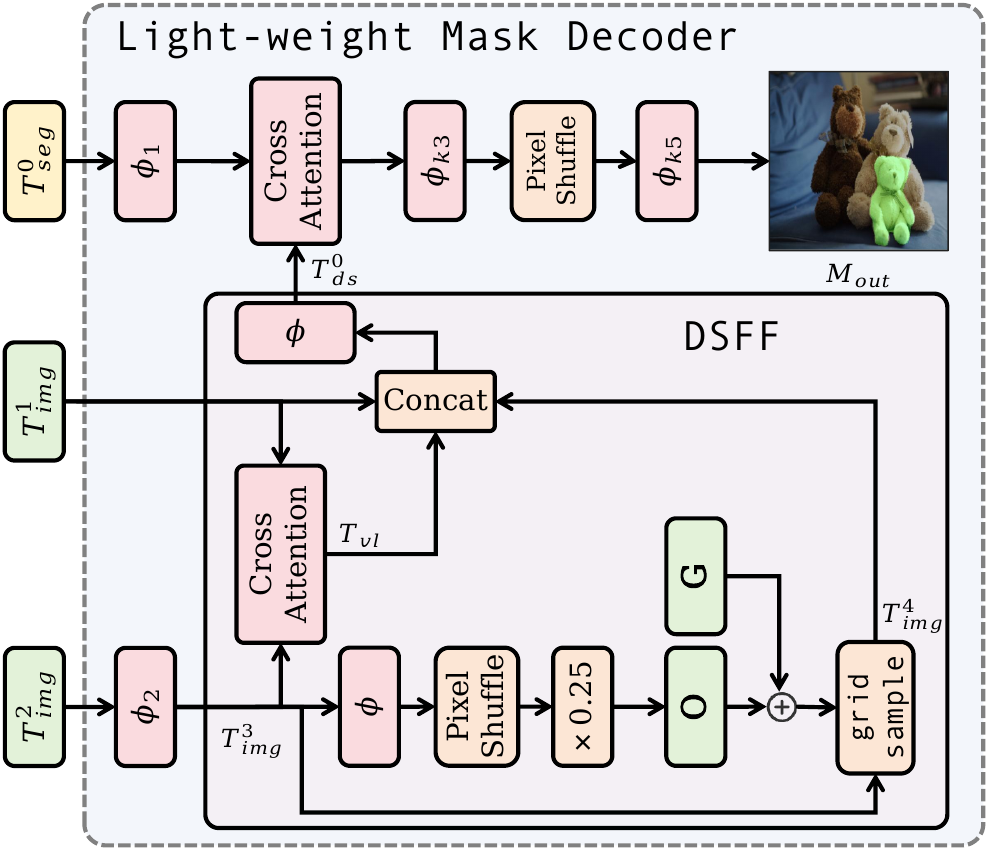} 
    \caption{Overview of the light-weight mask decoder with detail-enhanced and semantic-consistent feature fusion module (DSFF).}
    \label{fig:dsff2}  
\end{figure}

\begin{figure}[t] 
    \centering  
    \includegraphics[width=1.0\linewidth]{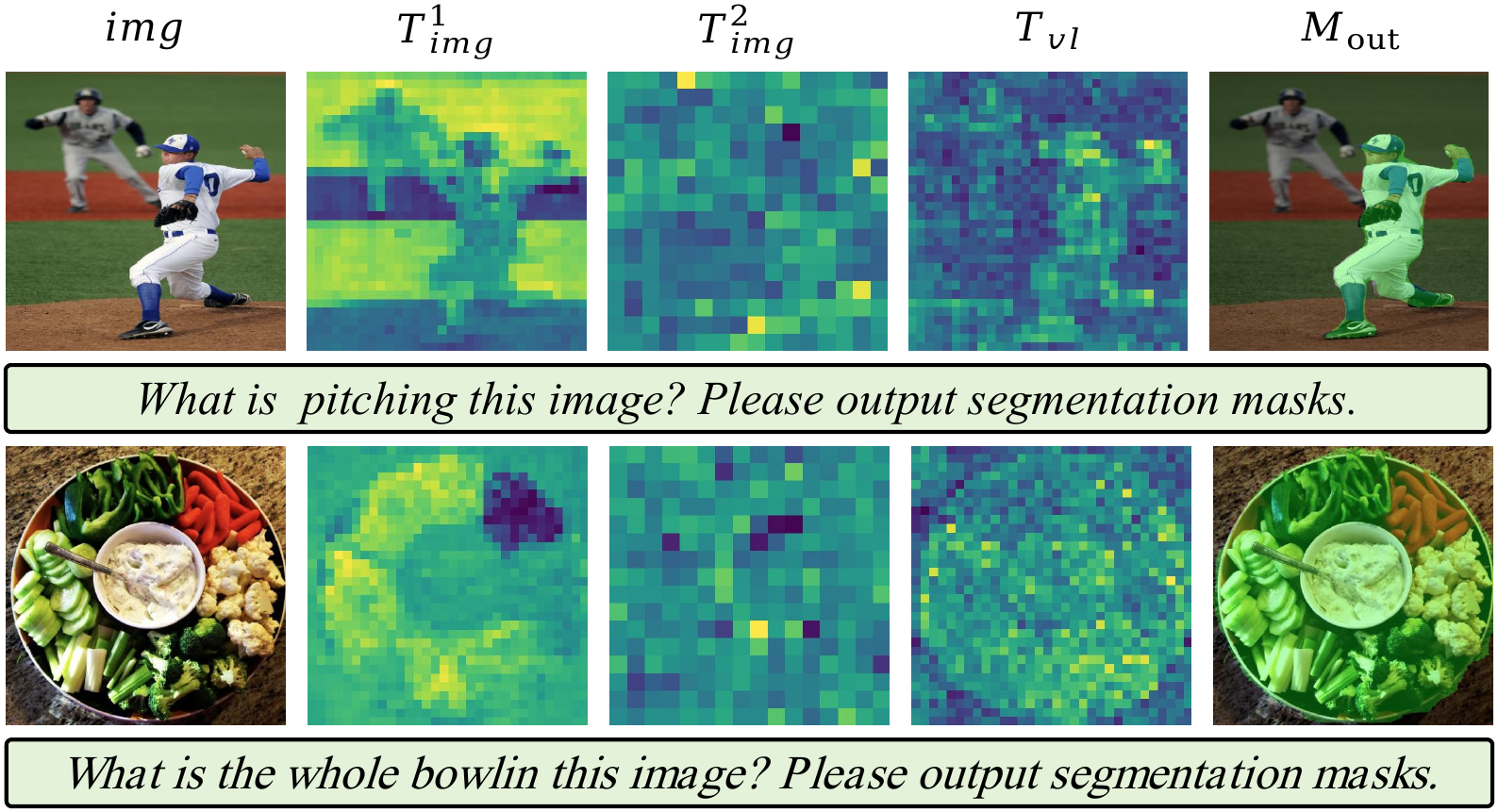} 
    \caption{Visualization of  $T^{1}_{img}$ from the visual encoder, $T^{2}_{img}$ from the large language model, and $T_{vl}$ from the cross-attention of $T^{1}_{img}$ and $T^{2}_{img}$.}
    \label{fig:dsff}  
\end{figure}

\subsubsection{Light-weight Mask Decoder.}
To adapt MLLMs to the RES task, the existing work, such as, LISA and GSVA, directly adopts the SAM for segmentation prediction, causing a large cost. To alleviate this issue, we carefully design a brand-new and novel light-weight mask decoder with Detail-enhanced and Semantic-consistent Feature Fusion module (DSFF), which optimally leverages detailed spatial features from the visual encoder and semantic features from the large language model to achieve precise mask prediction.

Specifically, for the light-weight design, as shown in Figure~\ref{fig:dsff2}, the semantic-related visual token $T^{2}_{img}$ and the special segmentation token $T^{0}_{seg}$ are first compressed through a linear projection, expressed as:
\begin{align}
    T^{3}_{img} = \phi_{1}(T^{2}_{img}), \quad
    T^{1}_{seg} = \phi_{2}(T^{0}_{seg}),
\end{align}
where $T^{3}_{img}$, $T^{1}_{seg}$, and $\phi(\cdot)$ denotes a linear layer for reducing the number of channels of features, increasing the information density, and reducing the computational complexity.

Inspired by the fact that the shallow visual feature $T^{1}_{img}$ extracted by the visual encoder contains rich details which are helpful for fine-grained segmentation, we specifically design a Semantic-consistent and Detail-enhanced Feature Fusion (DSFF) module to fuse the detail-related visual tokens $T^{1}_{img}$ and the compressed semantic-related visual tokens $T^{3}_{img}$, and then we can obtain the refined visual feature $T^{0}_{ds}$ as:
\begin{equation}\label{eq:dsff}
    T^{0}_{ds} = \text{DSFF}(T^{1}_{img}, T^{3}_{img}).
\end{equation}
where $\text{DSFF}(\cdot)$ will be described in detail in Section~\ref{sec:sdff}. Attributed to our DSFF, $T^{0}_{ds}$ is rich in details and consistent with the semantics of natural language expression. The merit will be validated in Figure~\ref{fig:dsff} and Table~\ref{tab:abldsff} below.

In order to inject the fused visual feature $T^{0}_{ds}$ into the segmentation token $T^{1}_{seg}$ for boosting the mask prediction, we introduce a cross-attention mechanism, formulated as:
\begin{equation}
\begin{split}
    & T^{1}_{ds} = \phi_{out}(\text{Softmax}(\frac{QK^\mathrm{T}}{\sqrt{D}})V), \\
    {Q} & = \phi_Q({{T^{0}_{ds}}}), 
    {K} = \phi_K({{T^{1}_{seg}}}), 
    {V} = \phi_V({{T^{1}_{seg}}}),
\end{split}
\end{equation}
where $\phi_Q(\cdot)$, $\phi_K(\cdot)$, and $\phi_V(\cdot)$ are linear projection operations for obtaining the query, key, and value, respectively. $\phi_{out}(\cdot)$ represents a linear projection layer.


Then by subsequently passing $T^{1}_{ds}$ to a series of operations, we can get the final mask $M_{out}$, expressed as:
\begin{equation}
M_{out}=\phi_{k5}(\text{PixelShuffle}(\phi_{k3}(T^{1}_{ds}))), 
\end{equation}
where $\text{PixelShuffle}(\cdot)$ denotes the pixel shuffle layer~\cite{shi2016real}, $\phi_{k3}(\cdot)$ denotes a convolutional layer with kernel size $3\times3$, and $\phi_{k5}(\cdot)$ denotes a convolutional layer with kernel size $5\times 5$. Please note that in the visual encoder $F_{en}$, the size of the original image is reduced from $448 \times 448$ to $32 \times 32$, and a visual token represents a $14 \times 14$ patch.

\subsection{Detail-Enhanced and Semantic-Consistent Feature Fusion}
\label{sec:sdff}

In this subsection, we will provide the detailed design about the network module $\text{DSFF}(\cdot)$ in Eq.~\eqref{eq:dsff}.


\input{Tables/RES}

Considering that the shallow detail-related visual feature $T_{img}^{1}$ is independently extracted by the visual encoder, to enhance features pertinent to the user's prompt, we fuse visual detail and semantic features via cross-attention, using $T_{img}^{1}$ as the Query and $T_{img}^{3}$ as the Key and Value, as:
\begin{equation}
\begin{split}
    & T_{vl} = \phi_{vl}(\text{Softmax}(\frac{QK^\mathrm{T}}{\sqrt{D}})V), \\
    {Q} & = \phi_Q({{T^{1}_{img}}}), 
    {K} = \phi_K({{T^{3}_{img}}}), 
    {V} = \phi_V({{T^{3}_{img}}}).
\end{split}
\end{equation}
As validated in Figure \ref{fig:dsff}, the shallow feature $T_{{img}}^{1}$ presents more details, but is not related to the text description, due to the non-directional nature of the visual encoder during feature extraction.
Additionally, the feature $T_{{img}}^{3}$ output by the LLM lacks fine-grained details, and the abstract semantic makes visual interpretation challenging. However, after the cross-attention fusion procedure, the modelled feature $T_{{vl}}$ is highly relevant to the user's prompt, which also validates the role of the introduction of $T_{{img}}^{1}$.

Concurrently, we expect to further unleash the power of $T_{{img}}^{1}$ and enable the detail-related information $T_{{img}}^{1}$ to further boost the segmentation. To this end, apart from the cross-attention operation, we spatially upsample $T_{img}^{3}$ from $16 \times 16$ to $32 \times 32$ to make its size align with the size of $T_{img}^{1}$ for the concatenation fusion. Specifically, 
inspired by DySample~\cite{liu2023learning}, we first feed $T_{img}^{3}$ with the size of $1024 \times 16 \times 16$ to a linear layer and a pixel shuffling layer in order to generate an offset tensor $\mathbf{O} \in \mathbb{R}^{(2 \times 4) \times 16 \times 16}$, where the value $2$ in the first dimension denotes the $x$ and $y$ coordinate offsets, as:
\begin{equation}
    \mathbf{O} = \text{PixelShuffle}(\phi(T^{3}_{img})),
\end{equation}
In order to avoid aliasing and overlapping between neighboring sampling points, we apply a fixed scope factor $\alpha = 0.25$ to constrain the offset magnitude:
\begin{equation}
    \mathbf{S} = \mathbf{G}+\alpha\cdot\mathbf{O},
\end{equation}
where $\mathbf{G}$ is the original sampling grid. The high-resolution output $T^{4}_{img}$ is obtained by sampling the input $T^{3}_{img}$ at the dynamically predicted positions $\mathbf{S}$ using the built-in differentiable grid\_sample function in PyTorch:
\begin{equation}
    T^{4}_{img} = \text{grid\_sample}(T^{3}_{img}, \mathbf{S}),
\end{equation}
In order to retain as much valid information as possible, we concatenate $T^{1}_{img}$, $T^{4}_{img}$, and $T_{vl}$ in the channel dimension, and then compress the channels using a linear layer to obtain $T^0_{ds}$ in Eq.~\eqref{eq:dsff}, as:
\begin{equation}
    T^0_{ds} = \phi(\text{Concat}(T^{1}_{img}, T^{4}_{img}, T_{vl})).
\end{equation}

\noindent {\textbf{Remark:}} As seen, through the cross-attention mechanism and the upsampled-based alignment procedure, we fully inject the shallow detail-related feature extracted by the visual encoder into the output of LLM. It is this way of fully utilizing features that makes it possible to design a lightweight mask decoder. Therefore, even if we do not adopt heavy SAM, our method is still expected to achieve satisfactory performance, which will be verified by experiments.


\input{Tables/REC}
\input{Tables/GRES}

\subsection{Training Objectives}
The model is trained end-to-end using the text generation loss $\mathcal{L}_{text}$ and the segmentation loss $\mathcal{L}_{mask}$, defined as:
\begin{equation}
\begin{split}
    &\mathcal{L}_{text} = \text{CE}(T^{llm}_{output}, T^{llm}_{gt}), \\
    &   \mathcal{L}_{mask} =  \text{BCE}(M_{out}, M_{gt})
    + \text{DICE}(M_{out}, M_{gt}),
    \end{split}
\end{equation}
where \text{CE} is Cross-Entropy, $\text{BCE}$ is per-pixel binary cross-entropy (BCE)~\cite{lai2024lisa}, $\text{DICE}$ is the DICE~\cite{rezatofighi2019generalized}, $T^{llm}_{gt}$ and $M_{gt}$ are the ground-truth. The total training loss is:
\begin{equation}
    \mathcal{L} = \lambda_{text} \mathcal{L}_{text} + \lambda_{mask} \mathcal{L}_{mask},
\end{equation}
where $\lambda_{text} = 1.0$ and $\lambda_{mask} = 1.0$.

\section{Experiments}

\subsection{Settings}
\subsubsection{Datasets.}
Follow previous works~\cite{text4seg}, we train on RefCOCO series~\cite{kazemzadeh2014referitgame, mao2016generation} for RES and REC tasks, the train split of gRefCOCO\cite{liu2023gres} for the GRES task. Crucially, unlike prior works (e.g., LISA \cite{lai2024lisa}, GSVA \cite{xia2024gsva}) that require extensive mixed-dataset pretraining followed by fine-tuning on RefCOCO-family datasets, our approach achieves state-of-the-art performance without any pretraining phase.  

\subsubsection{Evaluation Metrics.}
For RES, we use cIoU~\cite{xia2024gsva} as the evaluation metric, which computes the overall intersection over union across the dataset. 
For REC, as in previous work~\cite{lai2024lisa}, we consider the prediction correct if IoU is greater than 0.5 (Prec@0.5).
For GRES, we use both gIoU and cIoU metrics. In line with the setup from GRES~\cite{liu2023gres}, and gIoU measures the mean IoU per instance. To handle expressions that refer to no target, we employ No-target-accuracy (N-acc.)\cite{liu2023gres}, defined as the proportion of correctly identified empty-target cases. In such cases, a correct prediction is assigned a gIoU of 1.0 and is excluded from cIoU computation; incorrect predictions receive a gIoU of 0.0 and contribute to the union in cIoU.

\subsubsection{Implementation Details.}
In this paper, we use InternVL2-8B~\cite{chen2024far} and InternVL2.5-8B~\cite{internvl2.5} as our foundation models.
In training, we use a batch size of 32, running on 4 NVIDIA A100 GPUs for 50k iterations. 
The multimodal large language model we use is InternVL-8B.
The AdamW optimizer and a cosine annealing schedule are employed, with a learning rate of 4e-5. 
For efficient training, we employ LoRA with a rank of 64, freezing the visual encoder and LLM while keeping only the lora parameter of LLM and the mask decoder trainable. 
For RES tasks, the entire training process takes about 20 hours. 
After RES, we continued to fine-tune the model on the GRES task for 20k iterations, which took roughly 8 hours.
Further experimental details can be found in Appendix A.

\subsection{Main Results}

\input{Tables/ABL_DSFF}
\input{Tables/ABL_InternVL}

\subsubsection{Referring Expression Segmentation.}
To comprehensively evaluate {\mymethod}'s capabilities, we conduct rigorous benchmarking on standard RES tasks using RefCOCO, RefCOCO+, and RefCOCOg datasets, following established protocols: UNC \cite{kazemzadeh2014referitgame} partitions for RefCOCO/RefCOCO+ and UMD \cite{mao2016generation} split for RefCOCOg. As shown in Table~\ref{tab:res}, {\mymethod}${_\text{InternVL2}}$ achieves significant performance gains—outperforming Text4Seg${_\text{InternVL2}}$ with SAM by 1.3\% cIoU and surpassing SAM-free UFO${_\text{InternVL2}}$(only use features from the LLM output of MLLM for mask prediction) by 2.1\% cIoU. Furthermore, our {\mymethod}${_\text{InternVL2.5}}$ variant establishes state-of-the-art leadership across all RES benchmarks. Appendix B shows more qualitative results.

\subsubsection{Referring Expression Comprehension.}
    
    Table~\ref{tab:rec} shown the REC results of {\mymethod}. 
    Compared to Text4Seg$_{\text{InternVL2}}$ with SAM, {\mymethod}$_{\text{InternVL2}}$ achieved a 3.9\% improvement in Prec@0.5.
    Compared to HiMTok based on InternVL2.5~\cite{internvl2.5}, {\mymethod}$_{\text{InternVL2}.5}$ achieved a 1.6\% improvement in Prec@0.5.
    Compared to the currently widely used multimodal large models (IntrernvL2.5, Qwen2.5-VL), MLLMSeg achieves a performance improvement of over 3\%.
    In summary, our {\mymethod}$_{\text{InternVL2}.5}$ achieves overall leadership in REC.

\subsubsection{Generalized Referring Expression Segmentation.}
    To evaluate the performance of {\mymethod} on the GRES task, we utilize the gRefCOCO dataset~\cite{liu2023gres}. The dataset is partitioned into Train, Val, Test-A, and Test-B subsets, following the UNC split convention from RefCOCO~\cite{yu2016modeling}. Evaluation is conducted on the Val, Test-A, and Test-B subsets. 
    Table~\ref{tab:gres} shows the GRES results of {\mymethod}. 
    Compared with LISA and GSVA with SAM, our method achieves significant improvements in gIoU, cIoU, and recognition accuracy for empty targets.
    Compared with Text4Seg based on InternVL2, our approach also achieved significant leadership in gIoU and cIoU.

\subsection{Ablation Study}
\subsubsection{Effectiveness of DSFF.}
To validate the effectivess of our DSFF, we conducted experiments based on InternVL2.5~\cite{internvl2.5}. Specifically, we evaluated models utilizing image features from three configurations: solely the visual encoder, solely the LLM, and jointly from both the visual encoder and the LLM.
The results, reported in Table~\ref{tab:abldsff}, demonstrate that using either $T^{1}_{img}$ (from the visual encoder) or $T^{2}_{img}$ (from the LLM) in isolation for mask prediction yields comparable performance on the RES and REC tasks.
However, combining features from both sources allows the detailed and semantic information to complement each other, resulting in a measurable performance gain.
Crucially, upon integrating DSFF, the detailed and semantic features are effectively fused, leading to further performance improvements.
Remarkably, with 34M parameters, our approach achieves performance comparable to or exceeding that of MLLMs equipped with SAM (632M parameters).
These experimental results conclusively demonstrate the effectiveness of our DSFF.


\subsubsection{Performance at different model scales.}
To evaluate the effectiveness of our proposed lightweight mask decoder across MLLMs of varying scales, we conducted experiments using InternVL2.5 with 1B, 2B, and 4B base models.
As shown in Table~\ref{tab:abl_internvl}, while performance on the RES and REC tasks exhibits a modest degradation when reducing the model size from 8B to 4B, the overall accuracy remains consistently high. Critically, even with further reductions to 2B and 1B parameter scales, the model maintains acceptable performance without suffering catastrophic performance drops, demonstrating robustness against significant parameter reduction.
These results underscore the stability of our proposed method and highlight its suitability for deployment in resource-constrained scenarios.

\section{Conclusion}
In this work, to finely adapt MLLMs to deal with the RES task, we specifically propose {\mymethod}, a novel framework that fully utilizes the inherent fine-grained cues within MLLMs for RES. The proposed detail enhancement and semantically consistent feature fusion module, coupled with a light-weight mask decoder, enables effective integration of semantic understanding from LLM with high-resolution visual cues extracted by the inherent visual encoder.  Experiments demonstrate that {\mymethod} achieves competing performance with only 34M network parameters, establishing a new paradigm for efficient and accurate RES. 


\bibliography{aaai2026}

\end{document}

%% file: Tables/RES.tex
\begin{table*}[h]
\centering
    \begin{tabular}{lc|ccc|ccc|cc|c}
    \hline
    \multirow{2}*{Methods}  & \multirow{2}*{w/ SAM} & \multicolumn{3}{c|}{RefCOCO} & \multicolumn{3}{c|}{RefCOCO+} & \multicolumn{2}{c|}{RefCOCOg} & \multirow{2}*{Avg.}\\
    \cline{3-10}
     & & Val & TestA & TestB & Val & TestA & TestB & Val & Test & \\
    \hline
    LISA(ft)\cite{lai2024lisa}          & \ding{51} & 74.9 & 79.1 & 72.3 & 65.1 & 70.8 & 58.1 & 67.9 & 70.6 & 69.9 \\
    GSVA(ft)\cite{xia2024gsva}          & \ding{51} & 77.2 & 78.9 & 73.5 & 65.9 & 69.6 & 59.8 & 72.7 & 73.3 & 71.4 \\
    AnyRef(ft)\cite{he2024multi}     & \ding{51} & 76.9 & 79.9 & 74.2 & 70.3 & 73.5 & 61.8 & 70.0 & 70.7 & 72.2 \\
    PixelLM\cite{ren2024pixellm}    & \ding{55} & 73.0 &  76.5 &  68.2 &  66.3 &  71.7 &  58.3 &  69.3 &  70.5 & 69.2  \\
    VisionLLM v2 \cite{wu2024visionllm} & \ding{55} & 76.6 & 79.3 & 74.3 & 64.5 & 69.8 & 61.5 & 70.7 & 71.2 & 71.0 \\
    Text4Seg$_{\text{InternVL2}}$\cite{text4seg}  & \ding{55} & 74.7 & 77.4 & 71.6 & 68.5 & 73.6 & 62.9 & 70.7 & 71.6 & 71.4 \\
    Text4Seg$_{\text{InternVL2}}$\cite{text4seg}  & \ding{51} & 79.2 & 81.7 & 75.6 & 72.8 & 77.9 & 66.5 & 74.0 & 75.3 & 75.4  \\
    M²SA\cite{jang2025mmr} & \ding{51} & 74.0 & 76.8 & 69.7 & 63.1 & 67.2 & 56.1 & 67.0 & 68.3 & 67.8 \\
    SegAgent$_{\text{LLaVA}}$\cite{zhu2025segagent}  & \ding{51}   & 79.2 & 81.4 & 75.7 & 71.5 & 76.7 & 65.4 & 74.8 & 74.9  &75.0 \\
    SegAgent$_{\text{Qwen}}$ \cite{zhu2025segagent}  & \ding{51}   & 78.0 & 80.3 & 75.0 & 70.9 & 75.5 & 65.8 & 74.5 & 74.6 & 74.3\\
    UFO$_{\text{InternVL2}}$\cite{tang2025ufo}  & \ding{51}  & 78.0 & 79.7 & 75.6 & 72.3 & 76.8 & 66.6 & 73.7 & 74.3 & 74.6   \\
    VRS-HQ \cite{gong2025devil}  & \ding{51}                 & 73.5 & 77.5 & 69.5 & 61.7 & 67.6 & 54.3 & 66.7 & 67.5 & 67.3 \\
    $\mymethod_{\text{InternVL2}}$ (Ours)      & \ding{55}   & 79.2  & 81.0  & 77.5  & 73.9  & 77.6 & 68.6 & 77.3 & 78.6 & 76.7 \\
    $\mymethod_{\text{InternVL2}.5}$ (Ours)    & \ding{55}   & \textbf{81.0}  & \textbf{82.4}  & \textbf{78.7} & \textbf{76.4} & \textbf{79.1} & \textbf{72.5} & \textbf{79.9} & \textbf{80.8} & \textbf{78.9} \\
    \hline
    \end{tabular}
    \caption{\textbf{Referring Expression Segmentation} results (cIoU) on RefCOCO (+/g) datasets. The models we compared are all MLLMs ($\leq$8B) based methods and the best results are marked in bold.} 
    \label{tab:res}
\end{table*}

%% file: Tables/REC.tex
\begin{table*}[h]
\centering
    \begin{tabular}{lc|ccc|ccc|cc|c}
    \hline
    \multirow{2}*{Methods}  & \multirow{2}*{w/ SAM} & \multicolumn{3}{c|}{RefCOCO} & \multicolumn{3}{c|}{RefCOCO+} & \multicolumn{2}{c|}{RefCOCOg} & \multirow{2}*{Avg.}\\
    \cline{3-10}
     & & Val & TestA & TestB & Val & TestA & TestB & Val & Test & \\
    \hline

    
    Qwen2-VL\cite{Qwen2VL}   & \ding{55} & 91.7 &  93.6 &  87.3 & 85.8 & 90.5 &  79.5 & 87.3 & 87.8 & 87.9 \\
    InternVL2\cite{chen2024far}  & \ding{55} & 87.1 & 91.1 & 80.7 & 79.8 & 87.9 & 71.4 & 82.7 & 82.7 & 82.9 \\
    LISA(ft)\cite{lai2024lisa}  & \ding{51} & 85.4 & 88.8 & 82.6 & 74.2 & 79.5 & 68.4 & 79.3 & 80.4 & 79.8 \\
    GSVA(ft)\cite{xia2024gsva}  & \ding{51} & 86.3 & 89.2 & 83.8 & 72.8 & 78.8 & 68.0 & 81.6 & 81.8 & 80.3 \\
    PixelLM\cite{ren2024pixellm} & \ding{55} & 89.8 & 92.2 & 86.4 & 83.2 & 87.0 & 78.9 & 84.6 & 86.0 & 86.0\\
    VisionLLM v2 \cite{wu2024visionllm}  & \ding{55} & 87.9 & 91.2 & 84.3 & 77.6 & 83.8 & 70.2 & 82.9 & 84.1 & 82.8 \\
    InternVL2.5\cite{internvl2.5} & \ding{55} &  90.3 & 94.5 & 85.9 & 85.2 & 91.5 & 78.8 & 86.7 & 87.6 &  87.6 \\
    Qwen2.5-VL\cite{bai2025qwen2.5}   & \ding{55} &  90.0 &  92.5 &   85.4 & 84.2 &  89.1 &  76.9 &  87.2 &  87.2 & 86.6 \\
    Text4Seg$_{\text{InternVL2}}$\cite{text4seg}  & \ding{55} & 88.3 & 91.4 & 85.8 & 83.5 & 88.2 & 77.9 & 82.4 & 82.5 & 85.0 \\
    Text4Seg$_{\text{InternVL2}}$\cite{text4seg}  & \ding{51} & 90.3 & 93.4 & 87.5 & 85.2 & 89.9 & 79.5 & 85.4 & 85.4 & 87.1\\
    UFO$_{\text{InternVL2}}$\cite{tang2025ufo}  & \ding{55} & 91.4  & 93.8  & 88.2  & 85.7  & 90.7  & 79.7  & 86.8  & 87.4  & 88.0 \\
    HiMTok$_{\text{InternVL2}.5}$\cite{wang2025himtok} & \ding{55} & 92.9 &  94.7 &  89.3 &  87.6 &  91.5 &  81.5 &  88.5 &  89.0 & 89.4 \\
    $\mymethod_{\text{InternVL2}}$ (Ours)         & \ding{55} & 92.0 & 94.0 & 89.0 & 87.4 & 90.7 & 81.8 & 89.0 & 88.9 & 89.1 \\
    $\mymethod_{\text{InternVL2}.5}$ (Ours)       & \ding{55} & \textbf{93.5}  & \textbf{95.0}  & \textbf{90.3}  & \textbf{89.4}  & \textbf{92.3}  & \textbf{85.2}  & \textbf{91.1}  & \textbf{91.3}  & \textbf{91.0}  \\
    \hline
    \end{tabular}
    \caption{\textbf{Referring Expression Comprehension} results (Prec@0.5) on RefCOCO (+/g) datasets. The models we compared are all MLLMs ($\leq$8B) based methods and the best results are marked in bold.} 
    \label{tab:rec}
\end{table*}

%% file: Tables/GRES.tex
\begin{table*}[h]
    \centering
    \begin{tabular}{l|c|ccc|ccc|ccc}
    \hline
    \multirow{2}*{Methods}  & \multirow{2}*{w/ SAM} & \multicolumn{3}{c|}{Validation Set} & \multicolumn{3}{c|}{Test Set A} & \multicolumn{3}{c}{Test Set B} \\
    \cline{3-11}
     & & gIoU & cIoU & N-acc & gIoU & cIoU & N-acc & gIoU & cIoU & N-acc \\
    \hline
    LISA\cite{lai2024lisa}         & \ding{51} & 32.2 & 38.7 & 2.71   & 48.5 & 52.6 & 6.37   & 39.7 & 44.8 & 5.0 \\
    LISA(ft)\cite{lai2024lisa}    & \ding{51} & 61.6 & 61.8 & 54.7   & 66.3 & 68.5 & 50.0   & 58.8 & 60.6 & 51.9 \\
    GSVA\cite{xia2024gsva}         & \ding{51} & 63.3 & 61.7 & 56.5   & 70.1 & 69.2 & 63.5   & 61.3 & 60.3 & 58.4 \\
    GSVA(ft)\cite{xia2024gsva}     & \ding{51} & 66.5 & 63.3 & 62.4   & 71.1 & 69.9 & 65.3   & 62.2 & 60.5 & 60.6 \\
    LaSagnA* \cite{wei2024lasagna}  & \ding{51} &  32.4   & 38.1 & - &  47.3  &  50.4 & - &  38.9  &  42.1 &  - \\
    PSALM*\cite{zhang2024psalm}       & \ding{51} & 43.3  & 42.0  & -&  54.5 &  52.4  & -  &  52.5  & 50.6 & -  \\
    SAM4MLLM\cite{chen2024sam4mllm}  & \ding{51}  & 71.9 &  67.8 & - & 74.2 & 72.2 & - & 65.3 & 63.4 & - \\
    Text4Seg$_{\text{InternVL2}}$\cite{text4seg}  & \ding{55} & 71.8 & 65.6 & -   & 71.2 & 70.0 & -   & 64.2 & 62.5 & - \\
    Text4Seg$_{\text{InternVL2}}$\cite{text4seg}  & \ding{51}  & 74.4 & 69.1 & -   & 75.1 & 73.8 & -   & 67.3 & 66.6 & - \\
    \mymethod$_{\text{InternVL2}}$ (Ours)   & \ding{55} & 73.2	& 70.0 & 71.1 & 75.2 & 75.4 & 68.5 & 67.9 & 67.1 & 63.1 \\
    \mymethod$_{\text{InternVL2}.5}$ (Ours) & \ding{55} & \textbf{75.1} & \textbf{71.6} & \textbf{73.2} & \textbf{77.0} & \textbf{76.9} & \textbf{72.4} & \textbf{69.7} & \textbf{68.5} & \textbf{65.5}  \\
    \hline
    \end{tabular}
    \caption{\textbf{Generalized Referring Expression Segmentation} results on the gRefCOCO dataset. * indicates zero-shot performance. The models we compared are all MLLMs ($\leq$8B) based methods, and the best results are marked in bold.}
    \label{tab:gres}
\end{table*}

%% file: Tables/ABL_DSFF.tex
\begin{table*}[h]
    \centering
    \begin{tabular}{l|c|c|ccc|ccc|cc|c}
    \hline
    \multirow{2}*{Image Features} & {Feature} & {Mask Decoder }  & \multicolumn{3}{c|}{RefCOCO} & \multicolumn{3}{c|}{RefCOCO+} & \multicolumn{2}{c|}{RefCOCOg} & \multirow{2}*{Avg.}\\
    \cline{4-11}
      & Fusion & \#Params  & Val & TestA & TestB & Val & TestA & TestB & Val & Test & \\
    \hline
    \multicolumn{11}{c}{\textbf{Referring Expression Segmentation}} \\
    $T^{1}_{img}$  & None & 17.56M  & 79.7  & 81.6  & 77.4  & 75.1  & 78.8  & 70.4  & 78.5  & 79.4  & 77.6 \\
    $T^{2}_{img}$  & None & 21.78M  & 79.5  & 81.0  & 78.0  & 75.1  & 77.8  & 70.9  & 78.8  & 79.6  & 77.6 \\
    $T^{1}_{img}$, $T^{2}_{img}$  & Concat  & 22.83M  & 80.4  & 82.0  & 78.3  & 75.4  & 78.6  & 70.7  & 79.3  & 80.3  & 78.1 \\
    $T^{1}_{img}$, $T^{2}_{img}$  & DSFF    & 34.39M  & 81.0  & 82.4  & 78.7  & 76.4  & 79.1  & 72.5  & 79.9  & 80.8  & 78.9 \\
    \hline
    \multicolumn{11}{c}{\textbf{Referring Expression Comprehension}} \\
    $T^{1}_{img}$  & None & 17.56M  & 92.6 & 94.6 & 89.3 & 88.6 & 92.4 & 83.5 & 90.7 & 91.2 & 90.4  \\
    $T^{1}_{img}$  & None & 21.78M  & 92.8 & 94.4 & 90.5 & 88.7 & 91.7 & 84.4 & 90.3 & 90.5 & 90.4  \\
    $T^{1}_{img}$, $T^{2}_{img}$  & Concat  & 22.83M  & 93.1 & 95.1 & 90.3 & 88.8 & 92.6 & 83.2 & 90.9 & 90.6 & 90.6  \\
    $T^{1}_{img}$, $T^{2}_{img}$  & DSFF    & 34.39M  & 93.5 & 95.0 & 90.3 & 89.4 & 92.3 & 85.2 & 91.1 & 91.3 & 91.0  \\
    \hline
    \end{tabular}
    \caption{Performance comparison of different image feature fusion strategies on RES and REC across RefCOCO (+/g).} 
    \label{tab:abldsff}
\end{table*}


%% file: Tables/ABL_InternVL.tex
\begin{table*}[h]
    \centering
    \begin{tabular}{ll|ccc|ccc|cc|c}
    \hline
    \multirow{2}*{Models}  & \multirow{2}*{LLMs}  & \multicolumn{3}{c|}{RefCOCO} & \multicolumn{3}{c|}{RefCOCO+} & \multicolumn{2}{c|}{RefCOCOg} & \multirow{2}*{Avg.}\\
    \cline{3-10}
      &  & Val & TestA & TestB & Val & TestA & TestB & Val & Test & \\
    \hline
    \multicolumn{11}{c}{\textbf{Referring Expression Segmentation}} \\
    \mymethod$_{\text{InternVL-1B}}$  & Qwen2.5-0.5B-Instruct   &  77.3  & 79.9  & 74.9  & 71.4  & 78.4  & 66.4  & 75.1  & 75.9  & 74.9  \\
    \mymethod$_{\text{InternVL-2B}}$  & internlm2\_5-1\_8b-chat &  79.3  & 81.0  & 77.1  & 73.6  & 77.5  & 69.4  & 76.5  & 78.4  & 76.6  \\
    \mymethod$_{\text{InternVL-4B}}$  & Qwen2.5-3B-Instruct     &  79.5  & 81.4  & 77.7  & 75.3  & 78.8  & 70.8  & 78.1  & 78.4  & 77.5  \\
    \mymethod$_{\text{InternVL-8B}}$  & internlm2\_5-7b-chat    &  81.0  & 82.4  & 78.7  & 76.4  & 79.1  & 72.5  & 79.9  & 80.8  & 78.9  \\
    \hline
    \multicolumn{11}{c}{\textbf{Referring Expression Comprehension}} \\
    \mymethod$_{\text{InternVL-1B}}$  & Qwen2.5-0.5B-Instruct   &  89.1  & 92.1  & 85.9  & 83.4  & 88.7  & 78.5  & 87.4  & 88.1  & 86.7  \\
    \mymethod$_{\text{InternVL-2B}}$  & internlm2\_5-1\_8b-chat &  91.3  & 93.6  & 88.1  & 86.3  & 90.6  & 81.6  & 89.1  & 89.9  & 88.8  \\
    \mymethod$_{\text{InternVL-4B}}$  & Qwen2.5-3B-Instruct     &  92.1  & 93.9  & 88.8  & 87.9  & 91.5  & 82.6  & 89.4  & 89.8  & 89.5  \\
    \mymethod$_{\text{InternVL-8B}}$  & internlm2\_5-7b-chat    &  93.5  & 95.0  & 90.3  & 89.4  & 92.3  & 85.2  & 91.1  & 91.3  & 91.0  \\
    \hline
    \end{tabular}
    \caption{Performance comparison of different foundation MLLMs on RES and REC across RefCOCO (+/g).} 
    \label{tab:abl_internvl}
\end{table*}